\RequirePackage{fix-cm}
\documentclass[twocolumn]{svjour3}           %
\smartqed  %
\usepackage{graphicx}

\usepackage{picinpar}  
\begin{document}

\title{Defining Explainable AI for Requirements Analysis}

\author{Raymond Sheh \and
        Isaac Monteath }

\institute{Raymond Sheh and Isaac Monteath \at
	      (formerly) Department of Computing, Curtin University, Bentley WA 6102, Australia \\
              \email{ray@raymondsheh.org}           }

\date{}

\maketitle

\begin{abstract}

Explainable Artificial Intelligence (XAI) has become popular in the last few
	years. The Artificial Intelligence (AI) community in general, and the
	Machine Learning (ML) community in particular, is coming to the
	realisation that in many applications, for AI to be trusted, it must
	not only demonstrate good performance in its decisionmaking, but it
	also must explain these decisions and convince us that it is making the
	decisions for the right reasons. However, different applications have
	different requirements on the information required of the underlying AI
	system in order to convince us that it is worthy of our trust. How do
	we define these requirements? 

In this paper, we present three dimensions for categorising the explanatory
	requirements of different applications. These are Source, Depth and
	Scope. We focus on the problem of matching up the explanatory
	requirements of different applications with the capabilities of
	underlying ML techniques to provide them. We deliberately avoid
	including aspects of explanation that are already well-covered by the
	existing literature and we focus our discussion on ML although the
	principles apply to AI more broadly.

\end{abstract}

\section{Introduction}

Artificial Intelligence (AI) in general, and Machine Learning (ML) in
particular, have achieved impressive levels of performance in replicating
and exceeding human capabilities in several high profile decision-making and
behaviour generation tasks. Techniques such as Neural Networks and Deep
Learning \cite{schmidhuber2015deep} have always generated remarkable results,
ranging from work on self-driving cars \cite{pomerleau1989alvinn}
to beating human champions in games such as Go \cite{silver2016mastering}. 
However, much of this advancement has come at the expense of the ability for
humans to trace, understand, verify and learn from these intelligent systems.
This is particularly important in applications where systems must be traceable,
reliable or held accountable for safety or regulatory purposes. Indeed, while
good software engineering practice has resulted in software becoming more
modular, reliable, testable, predictable and accountable, machine learned
systems are becoming more tightly coupled and their interactions increasingly
obfuscated and opaque \cite{sculley14machine}. 

This increasing obfuscation of the decisionmaking processes of AI agents and systems 
will inevitably lead to an erosion of trust. The increasing number of incidents in the 
popular media of AI systems making incorrect decisions has only accelerated this 
erosion. To be trusted, a system has to demonstrate competence (the capability to 
perform the task, make the decision or provide the information), honesty (that the 
process leading to the decision is transparent and accountable) and alignment 
(that the agent or system doesn't have an ulterior motive and hasn't been 
compromised) \cite{freed2018explanation}. Explanations form a vital part of 
satisfying these requirements. 

Explainable AI (XAI) seeks to produce AI agents and systems that can not only make
decisions but can also appropriately satisfy the application's requirements for
associated explanations. Although it can trace its origins to the earliest days
of AI research, particularly in the Expert Systems community
\cite{Swartout1983,VanLent2001}, XAI has become a very topical area in the last
couple of years. 

To properly develop and deploy XAI capabilities to answer these requirements,
we need semantics that encapsulate these various capabilities and requirements
under one unified set of categories. This paper presents a survey of existing
categorisations and highlights where they are insufficient. We then present a novel, unified set of categories that we propose will
answer this need.

The contribution of our work is a set of categories that enable those who
specify, use, develop or otherwise rely on XAI systems to trade off the human
need for explainability against other factors such as efficiency, predictive
accuracy and representational flexibility of the underlying ML systems. Each
category also outlines the needs of the aforementioned users. These categories
are an evolution of those presented by \cite{sheh2017different} and complement
existing human factors and cognitive science work on what explanations are
necessary and how they should be presented. In a sense, the underlying concepts
are not new and exist implicitly and in parts within other classifications. Our
contribution is in bringing them together into an explicit, unified set of
semantics that can be used for requirements and capability analysis for the
purpose of trust in AI systems.

An infinite number of explanations can explain a given decision, just as an
infinite number of functions can explain a single datapoint. This is true even
ignoring factors of granularity, presentation and abstraction. 
Similarly, a subset of these explanations (although still possibly infinite)
will explain any finite set of decisions on actions that the system or agent
may make and only one explanation will truly match the system's or agent's
underlying decision process.

Much of the existing commentary on explainability, especially from the
cognitive science perspective, is accepting of the idea that an explanation can
be useful if it matches the decisions, without being some version of the ``one true''
explanation.  For example, \cite{Bibal2016} assert that understandability is
the most important problem, even beyond the source of the explanation (which they refer to as
``accuracy'').  This may be true when we have no way of knowing what the ``one
true'' explanation is. This is often the case for outward-looking explanations.						We may never truly know why a company went bankrupt although we may have
some good theories. Crucially, this is also often the case for human explanations
where so much of the decision-making process is not the result of conscious thought.

\section{Background}

\label{ssec:excat}

There is much existing work on defining and categorising explanations,
especially from the cognitive science perspective. 
Some, such as \cite{Keil2003Trends},
are concerned with how to focus explanations onto what is most
informative or desirable for a given audience and how this changes with
application. Other work such as \cite{Tolchinsky2012}
focus on how AI systems can use explanations in dialog to convincingly justify their decisions. \cite{Miller2017Explainable} discuss why it is important to ``beware of the inmates 
running the asylum'', noting that explanations should be
targeted towards the right user audience -- who are not necessarily the AI
practitioner, software engineer, or even the domain expert. This may 
be important from the perspective of presentation to the end user. However, 
we argue that it is precisely the AI practitioners, software engineers or 
domain experts who place greater demands on where the underlying information 
for explanations come from. Therefore, it is important to consider 
this as a factor in the discussion, separate from the presentation of the explanation itself.

With so many groups working on the topic of explainability, even the semantics
surrounding explainability, such as its definition, requirements and metrics, have
become overloaded.  Recent work brings together some of
these semantics but also highlights where they are deficient, particularly with
regard to the source and detail available in the underlying information.
This makes these properties difficult to evaluate in an intelligent system or agent.
\cite{Bibal2016} also make an important distinction between the terms
``interpretable model'', which requires an underlying ``white-box" model such
as a decision tree, and ``interpretable representation'' which refers to how any
underlying model is represented to the end-user. Although their paper
highlights many of the varied applications where the requirements are poorly
covered by existing semantics, they do not propose a unified set of categories
to discuss these at a level of fidelity necessary to properly consider
the trade-offs. 

\cite{Montavon2018} defines explanations as distinct from interpretations. An
explanation is ``...  the collection of features of the interpretable domain,
that have contributed for a given example to produce a decision ...'' while an
interpretation is ``... the mapping of an abstract concept (e.g. a predicted
class) into a domain that the human can make sense of.'' We feel that this
definition of explanation is unnecessarily narrow. There are many times
when we demand more of an explanation than simply the collection of features.
We want to know more about what the system knows in its decision-making.

According to \cite{Doshi-velez2017}, current techniques for evaluating
interpretability fall into two categories. The first proposes that if an AI agent or system
is useful in some practical application (or a simplified version of it), then
there must be a way to interpret the logic behind its actions. The second
involves using a model that is intrinsically
interpretable, and fitting it to solve a problem required by the application.
While the interpretability of both approaches can be tested
using human-beings, neither provides a rigorous framework for quantifying
interpretability.

\section{\label{sec:explanation}Categorising Explanations}

\begin{figure}
      \begin{center}
              \includegraphics[width=0.4\textwidth]{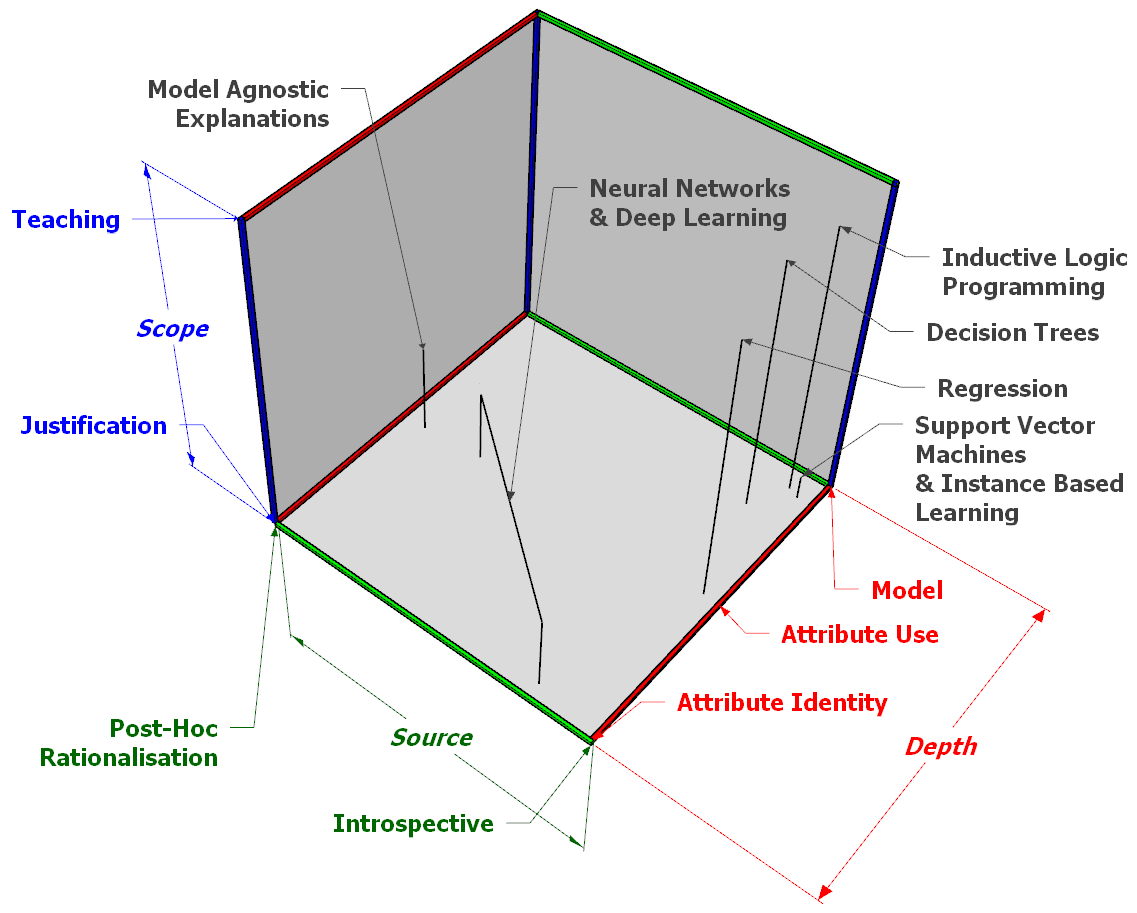}
      \end{center}

      \caption{\label{fig:examples}Three dimensions of explanation with
      examples of the explanatory capabilities possible using different 
      techniques. Axes are arbitrary. Note that each technique covers a
      ``cube'' that extends to the lower left corner but has been omitted for
      clarity. } 

\end{figure}

As introduced earlier,
we divide explanations along three
dimensions: \textbf{Source}, \textbf{Depth} and \textbf{Scope}. These are
visualised in Figure~\ref{fig:examples} along with the placement of some
examples of different AI techniques. In this section we present our definitions
for these dimensions. 

We consider the \textbf{Source} of the explanation to be where the explanatory
information came from. If it comes from another system or module, or from a
system observing a decision-making process as a black box and generating its
explanations with no more information than another observer might have, then we
consider the source to be \emph{Post-Hoc Rationalisation}. If instead the
explanatory information came from the same process that produced the underlying
decision and, crucially, retains symbolic meaning from that underlying decision
(as opposed to purely statistical information, for example), then we consider
the source to be \emph{Introspective}. While much of the discussion around XAI
would seem to tend towards one end of this category or the other, as we will
discuss, there is also a continuum between these two extremes. 

The distinction between \emph{Post-Hoc Rationalisation} and \emph{Introspective} explanations
appears in most categorisations of explanation. For example,
\cite{Swartout1993} refer to this as ``Fidelity'' and \cite{Bibal2016} as
``Accuracy'', while the ``Opaque'' category \cite{Doran2017} corresponds to
\emph{Post-Hoc Rationalisation} in our vernacular. The purely \emph{Introspective} explanation
corresponds to the ``one true'' explanation while others are \emph{Post-Hoc Rationalisation}. 

To the best of our knowledge, most XAI systems in the contemporary ML
literature tend to be \emph{Post-Hoc Rationalisation} systems of varying
degrees. This is perhaps due to the interest in Deep Learning, which is
necessarily opaque \cite{ribeiro2016why,Ross2017Right,Huang2017}.
\emph{Post-Hoc Rationalisation} Explanations may only be valid around a local
decision or some (perhaps often-visited or optimised) subset of the state
space, or else they may attempt to be globally valid. Just as in function
fitting, the nature of the deviation will depend on how the explanation is
generated. In the general case, there is no way to know the location or extent
of these deviations beyond probing the space, thereby limiting the ability of
\emph{Post-Hoc Rationalisation} Explanations to predict the system's decisions
beyond instances seen in operation or training. 

\emph{Introspective} Explanations are based on the system's underlying decision
process. As such, they are the accurate, ``true'' explanation for a given
decision, to a particular level of abstraction. Such AI techniques and models
are often referred to as being ``explainable'' or ``transparent''.

It is vital to know if an explanation is \emph{Post-Hoc Rationalisation} or \emph{Introspective}
when used for correcting faults, predicting behaviour in critical systems, or for
compliance and accountability. \emph{Post-Hoc Rationalisation} explanations may explain the observed behaviour but
have limited relevance to the underlying decision or fault. However, it may be impossible
for the end-user to tell the difference between these types of explanation.

In dividing explanations and techniques between \emph{Post-Hoc Rationalisation} and
\emph{Introspective}, it is important to determine what is meaningful in a
given context.  Every program that can run on a deterministic computer is
arguably capable of an ``\emph{Introspective}'' explanation. In the limiting
case, all one needs to do is to dump out the memory contents at every
instruction.  Moving up one level of abstraction, a conventional deep neural
network can also be considered ``\emph{Introspective}'' insofar as it is
possible to simply print out all of the neural network weights and activations.
For all but the most trivial problems, these ``Execution'' versions of
introspective explanations are neither understandable nor efficient enough to
be useful.

The \textbf{Depth} of an explanation can be \emph{Attribute} or \emph{Model}. 
An explanation that describes how attributes were used to make decisions within the context of the model are referred to as \emph{Attribute} explanations. Those that
also include how the model was generated from training data and background
knowledge are referred to as \emph{Model} explanations. 
For example, an image classification system may provide an \emph{Attribute}
explanation in the form of a saliency map. If it were then able to answer a
question about how it arrived at that saliency map, such as from training data or
background knowledge, it would be considered capable of providing
\emph{Model} explanations. 

We find that \emph{Model}
explanations are largely forgotten in the XAI literature. The vast
majority of techniques, discussions and semantics assume that explanations are
limited to explaining how an AI system makes use of the attributes being observed
to generate its current answer. 
This distinction is important because it speaks to the ability of a system to
assist humans in uncovering the reason for failures, both for troubleshooting and
accountability purposes. After all, if an image
recognition system on an autonomous military agent incorrectly recognises a turtle as a
weapon \cite{Athalye2017}, then not only would the humans in charge
be interested in knowing the \emph{Attribute} explanation but they would also
want the \emph{Model} explanation, which may suggest where else the agent will make
the same mistake and, ideally, how to fix or mitigate the problem. 

Within \emph{Attribute} explanations we also consider a sub-dimension,
which we define as \emph{Attribute Identity} vs \emph{Attribute Use} explanations. A
system that is only capable of \emph{Attribute Identity} explanations is able
to describe which attributes were used in making its decision but is unable to
explain, in a way that is meaningful to humans, how they were used. The
definition of explanation proposed in \cite{Montavon2018} only covers
\emph{Attribute Identity} explanations.  Examples of these explanations include
statistics on which attributes were most important and saliency maps. 

\emph{Attribute Use} explanations are able to explain how those attributes were
used. A decision tree or logic rule chain are examples of \emph{Attribute Use}
explanations. It is important to note that meaningfulness can be nebulous. In the general
case we consider an explanation to be \emph{Attribute Use} if the AI system is able
to incorporate information about decision boundaries into its explanation, be
it \emph{Introspective}ly (by looking at the underlying model) or in a
\emph{Post-Hoc Rationalisation} fashion (such as by sampling).

Explanation \textbf{Scope} can be \emph{Justification} or \emph{Teaching}. 
If the intent of the explanation is focused on a specific decision or group of
decisions, it is a \emph{Justification} Explanation. Otherwise, it is a
\emph{Teaching} Explanation. \emph{Teaching} Explanations tend to be
forgotten in the XAI literature. The work of \cite{Cotton2017} comes
closest in acknowledging that there are explanations for justifications.

Like the aforementioned \emph{Model} explanations, we find \emph{Teaching}
explanations to be largely ignored by the existing literature on explainability. 
However, they are an important part of forming trust with an AI system,
particularly in terms of its ability to properly generalise outside of examples
that have been seen. These explanations can take many forms. At one end they
can still be rooted in instances. For example, answers to the inquiry ``What
are examples of other instances that you consider to be similar to this?''
would be a \emph{Teaching} explanation. These other instances displayed may be
from the past (training or otherwise) or they may be synthetic instances,
generated by the model. 

Beyond these explanations are those that concern understanding the decision
boundaries. 
``How far would I
need to move this attribute to change your decision?'' might be such a question
that would be asked of a high reliability system. The complexity of the
resulting answer could be considerable, especially for models or
applications where decision boundaries are complex.  At the other end are more
abstract concepts. ``What are attributes that were significant in deciding on
this class?'' or even, ``Teach me about the differences between these
classes'', for instance.

\section{Discussion}

In this
section, we will discuss some salient examples, first of application
requirements and then of AI technique capabilities. The intent is to highlight
some issues in using these categorisations, rather than to
be comprehensive.

\subsection{Application: User Experience}

An AI system, such as a service robot, might be asked why they made a decision
or exhibited a behaviour that was surprising. The human has a model for the
robot's behaviour that is different from the robot's observed behaviour. The
role of explanation is to help the user to better understand the agent's
behaviour, to be more comfortable and to make better use of it by performing
``model reconciliation'' \cite{Chakraborti2016}. This goes both ways. For a
good user experience, sometimes the robot should change its behaviour to make
it fit the humans' model, to make it more ``explicable''. 

In our view, such an application does not require explanation to be
\emph{Introspective}. Indeed, it may be better to select an explanation that
the user is more likely to understand or accept over what is true, so long as
it explains the robot's behaviour with sufficient plausibility where it is
likely to be observed. In terms of explanation depth, we would suggest that
such explanations would be mainly \emph{Attribute Identity} or \emph{Use}.  The
exception may be in situations where the agent or system is expected to learn
on-the-fly.  A service robot being asked ``Why did you go the long way around
the building?'' and answering ``Because Fred told me to go that way last
week.'' is providing an explanation at the \emph{Model} \textbf{Depth}. The
combination of a \emph{Model} \textbf{Depth} with \emph{Post-Hoc
Rationalisation} \textbf{Source} might seem contradictory.  We consider such an
explanation as explaining the decision in a \emph{Post-Hoc Rationalisation}
manner given the whole corpus of data available to the system -- training, past
experience and current instance.  In contrast, an explanation at the
\emph{Attribute} \textbf{Depth} only considers the current instance and model.

\subsection{Application: Forensic and Compliance}

Forensic and Compliance applications range from public safety (such as
self-driving cars) to privacy and ethics, such as the European Union's
General Data Protection Regulation (GDPR) \cite{goodman16european}. In
our categorisation, the explanations demanded in these applications would have
a \textbf{Source} of \emph{Introspective} and a \textbf{Scope} of \emph{Model}.
The human users demand the true reason for an AI system's decisions (\textbf{Depth}
of \emph{Justification}) or some understanding as to what the AI system will do
more generally (\textbf{Depth} of \emph{Teaching}). In such an application, the
use of AI techniques that do not readily provide these capabilities should be
used with extreme caution as verification of their performance is going to be
at least in part dependent on statistical behaviour. Interestingly, the highest
demands placed on explanation in terms of information source by this
application could perhaps also be the least demanding from a human factors
perspective. The human users here are heavily invested in understanding the
AI system's decisions. 

\subsection{AI: Neural Networks}

There has been much recent discussion about explainability with regards to
Neural Networks and associated Deep Learning techniques. Some say they are not
explainable \cite{doyle2003review,roth2004explanations} while others claim that it is quite possible to generate
explanations from them \cite{ribeiro2016why,hendricks2016generating}. In our observation, much of this disagreement comes
from a disagreement in the types of explanation that are expected.  Some
sectors of the ML community do not consider any explanations beyond those that
we would categorise as \emph{Attribute Identity}, \emph{Justification}
explanations -- the definition by \cite{Montavon2018} as discussed in
Section~\ref{ssec:excat} being one example. By such a narrow definition, Neural
Networks can be made explainable.
Having said that, \cite{Montavon2018} does survey work that is able to move
towards more \emph{Introspective} and \emph{Attribute Use} explanations. 

In this direction, there has been much recent interest in tricking deep learning
based image recognition system into reporting arbitrary results through minor
perturbations in the image \cite{Su2017,Huang2017} or placing a separate object
in the scene \cite{Brown2017}. Approaches that seek to explain such systems by
highlighting saliency in the images \cite{Bach2015} are helpful to at least
detect and identify the problem. More generally, other techniques such as
\cite{ribeiro2016why} attempt to analyse the behaviour of the network in a
black-box fashion for the purpose of increasing trust in the system. In many
applications, such as finding similar images or products, especially where
there is no adversary, this is all one might expect.

Some performers on the DARPA XAI program
\cite{gunning16explainable} are generating explanations from deep learning by
leveraging information from within the network. This may be by inspecting the
network after it has been trained, such as by using input gradients
\cite{Ross2017Right}, or by incorporating the learning of explanations into the
learning of the network itself \cite{hendricks2016generating}. Through this, significant intermediate
representations and decision boundaries can be found.  Some of these
explanations are partially \emph{Introspective},
\emph{Attribute Use}, \emph{Justification} explanations. These explanations are
still not necessarily truly ``\emph{Introspective}'' insofar as there's no
guarantee that they are the result of the exact same network operation that
generated the decision. However, it at least opens up the possibility that they
may use the same intermediate representations, providing a 
partial grounding in the decision-making process.

\subsection{AI: Decision Trees and Rule Learners}

Decision Trees, Inductive Logic Programming, Ripple-Down Rules and other
representations that lend themselves to rule-based representations tend to be
thought of as explainable and transparent. They can yield explanations that
satisfy all eight octants of our classification scheme, including the most
``difficult'', being \emph{Introspective}, \emph{Model}, \emph{Teaching}
explanations. 

Decision trees can be instrumented in order to obtain some of this underlying
information. For example, metadata for \emph{Model} explanations can be
retained from the training process or inferred after the fact. Knowledge of
the information theoretic underpinnings of how the tree is learned also furnish
information that can be used to show which rules were most important, which
were best supported by the current instance, and which other rules from other
branches might have been relevant had the instance changed slightly.

One must be careful in claiming that such models are usefully
explainable just because they have a traceable, perhaps deterministic
path from training data to model to decision. If the underlying phenomena being
predicted cannot be meaningfully expressed by the model then any explanation
will be forced and limited in its usefulness, even if the predictive accuracy
is sufficient. For example, a decision tree can learn a concept in 2D space with a
decision boundary defined by $y=x$ but it does so by ``shattering'' this simple
concept into a multitude of If-Then-Else rules that are as informative to a
human as the weights in a Neural Network. In our categorisation, such a
decision tree does not support \emph{Introspective} or \emph{Teaching} explanations on these
attributes.

\section{Conclusion}

Before we can match an application's explanatory requirements with AI
capabilities, we need to define explanation. There have been many proposed
definitions in the literature, some salient examples of which we have surveyed.
However, we have found that they tend to be overly narrow and, generally,
focused on the (equally important problem of) presentation of the explanation
rather than the source. 
As AI becomes more a part of critical decision-making, society as a whole is
beginning to demand more reliability, transparency and accountability from AI
agents and systems. However, explainability is a trade-off, often occurring at the expense
of other factors such as performance or development effort.  
In this paper, we have presented a novel categorisation for explanations along
three dimensions of \textbf{Source}, \textbf{Depth} and \textbf{Scope}, and
discussed these in the context of several salient examples of applications and
AI techniques. We propose that these cover the full space of explanations and
can be used to match applications' explanatory requirements with AI
capabilities to satisfy them.

Semantics for defining and categorising explanation is just part of the 
challenge in matching explanatory requirements with AI capabilities. 
Another part of the challenge is extending these semantics to define how
``repairable'' an AI system is based on its underlying AI techniques. In some cases
it may well be acceptable to simply retrain on the misclassified examples.
However, in other cases it may be necessary to have verifiable proof that not
only has a particular instance of error been solved, but that all similar
potential errors have also been solved without ``catastrophically forgetting''
\cite{french1999catastrophic} what the AI system might have otherwise done
correctly. Certainly mechanical infrastructure -- vehicles, for instance -- are
held to this standard, it is only a matter of time before certain AI systems
are held to similar standards.  Augmenting our categories with the semantics to
describe these orthogonal requirements is the subject of future work. 
Finally, metrics for explanation apply differently to the different categories 
of explanation that we presented. Developing these metrics is also the subject of 
future work.

\bibliographystyle{spmpsci}
\bibliography{ijcai18}

\end{document}